\documentclass[10pt, a4paper]{article}
\usepackage{lrec}
\usepackage{multibib}
\newcites{languageresource}{Language Resources}
\usepackage{graphicx}
\usepackage{tabularx}
\usepackage{soul}
\usepackage{amssymb}

\usepackage{epstopdf}
\usepackage[latin1]{inputenc}

\usepackage{hyperref}
\usepackage{xstring}
\usepackage{caption}
\usepackage{subcaption}
\usepackage{alltt}
\usepackage{ctable}
\usepackage{multirow}
\usepackage{adjustbox}

\title{Evaluation of Croatian Word Embeddings\\ }

\name{Luká\v{s} Svoboda$^{1}$, Slobodan Beliga$^{2}$}

\address{1) Department of Computer Science and Engineering, University of West Bohemia\\ 
	Univerzitn\'{i} 22, 306 14 Plze\v{n}, Czech Republic\\
	  2)  Department of Informatics, University of Rijeka\\
         Radmile Matej\v{c}i\'{c} 2, 51000 Rijeka, Croatia \\
         svobikl@kiv.zcu.cz$^{1}$, sbeliga@uniri.hr$^{2}$\\
}

\abstract{
Croatian is poorly resourced and highly inflected language from Slavic language family. Nowadays, research is focusing mostly on English.
We created a new word analogy corpus based on the original English \emph{Word2vec} word analogy corpus and added some of the specific linguistic aspects from Croatian language. 
Next, we created Croatian WordSim353 and RG65 corpora for a basic evaluation of word similarities.
We compared created corpora on two popular word representation models, based on \emph{Word2Vec} tool and \emph{fastText} tool.\\
Models has been trained on 1.37B tokens training data corpus and tested on a new robust Croatian word analogy corpus. Results show that models are able to create meaningful word representation. This research has shown that free word order and the higher morphological complexity of Croatian language influences the quality of resulting word embeddings.
}

\begin{document}

\maketitleabstract

\section{Introduction}

Word representation based on Distributional Hypothesis \cite{Harris:1954}, commonly referred to as word embeddings, represent words as vectors of real numbers from high-dimensional space. The goal of such representations is to capture the syntactic and semantic relationship between words. 

It was shown that the word vectors can be sucessfully used in order to improve and/or simplify many NLP applications \cite{Collobert08aunified,DBLP:journals/corr/abs-1103-0398}. There are also NLP tasks, where word embeddings does not help much \cite{Andreas:2014}. 

Most of the work is focused on English. Recently the community has realized that the research should focus on other languages with rich morphology and different syntax \cite{berardi2015word,elrazzaz2017methodical,koper2015multilingual,DBLPSvobodaB16}, but there is still a little attention to languages from Slavic family. These languages are highly inflected and have a relatively free word order.
Since there are open questions related to the embeddings in the Slavic language family, in this paper, we will focus mainly on Croatian word embeddings, from the South Slavic language family. With the aim of expanding existing findings about Croatian word embeddings, we will: 
\begin{enumerate}
	\item Compare different word embeddings methods on Croatian language that is not deeply explored highly inflected language.
	\item For the purposes of the word embeddings experiments, we created three new datasets. Two basic word similarity corpora based on original WordSim353\cite{ws353} and RG65\cite{RubensteinGoodenough65} translated to Croatian. Except the similarity between words, we would like to explore other semantic and syntactic properties hidden in word embeddings. A new evaluation scheme based on word analogies were presented in \cite{mikolov2013efficient}.
	Based on this popular evaluation scheme, we have created a Croatian version of original Word2Vec analogy corpus in order to qualitatively compare the performance of different models. 
	\item Empirically compare the results obtained from the Croatian language to the results obtained from the language from the group of Indo-European language family (i.e. English - the most commonly studied language).
\end{enumerate}

Nowadays, word embeddings  are typically obtained as a product of training neural network-based language models. Language modeling is a classical NLP task of predicting the probability distribution over the "next" word. In these models a word embedding is a vector in $ \mathbb{R}^{n}	$, with the value of each dimension being a feature that weights the relation of the word with a "latent" aspect of the language. These features are jointly learned from plain unannotated text data. This principle is known as the \textit{Distributional Hypothesis} \cite{Harris:1954}. The direct implication of this hypothesis is that the word meaning is related to the context where it usually occurs and thus it is possible to compare the meanings of two words by statistical comparisons of their contexts. This implication was confirmed by empirical tests carried out on human groups in \cite{RubensteinGoodenough65,Charles2000}. 

There is a variety of datasets for evaluating semantic relatedness between English words, such as \emph{WordSimilarity-353} \cite{ws353}, \emph{Rubenstein and Goodenough (RG)} \cite{RubensteinGoodenough65}, \emph{Rare-words} \cite{Luong-etal:conll13:morpho}, \emph{Word pair similarity in context} \cite{Huang-2012}, and many others. \cite{mikolov2013efficient} reported that word vectors trained with a simplified neural language model \cite{bengio2006neural} encodes syntactic and semantic properties of language, which can be recovered directly from space through linear translations, to solve analogies such as:
$\vec{king} - \vec{man} = \vec{queen} - \vec{woman}  $.
Evaluation scheme based on word analogies were presented in \cite{mikolov2013efficient}. 

To the best of our knowledge, only small portion of recent studies attempted evaluating Croatian word embeddings. In  \cite{zuanovic2014experiments} authors translated small portion from English analogy corpus to Croatian to evaluate their Neural based model. However, this translation was only made for a total of 350 questions. 
There is only one analogy corpus representing Slavic language family - Czech word analogy corpus presented in \cite{DBLPSvobodaB16}. 

Many methods have been proposed to learn such word vector representations. One of the Neural Network based models for word vector representation which outperforms previous methods on word similarity tasks was introduced in \cite{Huang-2012}. Word embeddings methods implemented in tool \emph{Word2Vec} \cite{mikolov2013efficient} and GloVe \cite{pennington2014glove} significantly outperform other methods for word embeddings. Word vector representations made by these methods have been successfully adapted on variety of core NLP tasks. Recent library \emph{FastText} \cite{bojanowski2016enriching} tool is derived from Word2Vec and enriches word embeddings vectors with subword information. 

\section{Proposed corpora}
Original Word2Vec analogy  corpus is composed by 19,558 questions divided in two tested group : semantic and syntactic questions, e.g. king : man = woman : queen. Fourth word in question is typically the predicted one. 

Our Croatian analogy corpus has 115,085 question divided in the same manner as for English into two tested group: semantic and syntactic questions.\\

Semantic questions are divided into 9 categories, each having around 20 - 100 word question pairs. Combination of question pairs gives overall 36,880 semantics questions:  

\begin{itemize}
	\item[-] \texttt{capital-common-countries}: 
	This group consist of 23 the most common countries. These countries were adopted from original Word2Vec analogies and having highest number of occurrences in text between all languages. 
	\item[-] \texttt{chemical-elements}: Represents 119 pairs of chemical elements with their shortcut symbol (i.e. O - Oxygen).
	\item[-] \texttt{city-state}: Gives 20 regions (states) inside the Croatia and gives one of city example in such region.
	\item[-] \texttt{city-state-USA}: 67 pairs of cities and corresponding states in USA. This category is adopted from original English word analogy test.
	\item[-] \texttt{country-world}: 118 pairs of countries with main cities from all over the world. Translated from original Word2Vec analogies.
	\item[-] \texttt{currency-shortcut}: 20 pairs of state currencies with its shortcut name (i.e. Switzerland - CHF).
	\item[-] \texttt{currency}: 20 pairs of states with their currencies (i.e. Japan - yen). Translated from original EN analogy corpus.
	\item[-] \texttt{eu-cities-states}: 40 word pairs of states from EU and their corresponding main city (i.e. Belgium - Brussels).
	\item[-] \texttt{family}: 41 word pairs with family relation in masculine vs feminine form (i.e. brother - sister).
\end{itemize}

Syntactic part of corpus is divided into 14 categories, consisting of 78,205 questions: 
\begin{itemize}
	\item[-] \texttt{jobs}: This category is language-specific, consist of 109 pairs of job positions in masculine$\times$  feminine form.
	\item[-] \texttt{adjective-to-adverb}: 32 pairs of adjectives and its representatives in adverb form. 
	\item[-] \texttt{opposite}: 29 pairs of adjectives with its opposites. This category collects words from which is easy to make its opposites usually with preposition "un" or "in", respective preposition "ne" in Croatian (i.e. certain - uncertain). Adopted from original EN word analogies. 
	\item[-] \texttt{comparative}: 77 pairs of adjectives and its comparative form (i.e. good - better).
	\item[-] \texttt{superlative}: 77 pairs of adjectives and its superlative form. 
	\item[-] \texttt{nationality-man}: 84 pairs of states and humans representing its nationalities in masculine form. (i.e. Switzerland - Swiss).
	\item[-] \texttt{nationality-female}: 84 pairs of states and its nationalities in feminine form. This is language specific.
	\item[-] \texttt{past-tense}: 40 pairs of verbs and its past tense form. 
	\item[-] \texttt{plural}: 46 pairs of nouns and its plural form. 
	\item[-] \texttt{nouns-antonyms}: 100 pairs of nouns and its antonyms. 
	\item[-] \texttt{adjectives-antonyms}: Similar category to \emph{opposite}, it consists of 96 word pairs of adjectives and their antonyms. However, words are much more complex (i.e. good - bad).
	\item[-] \texttt{verbs-antonyms}: 51 pairs of verbs and its antonyms.
	\item[-] \texttt{verbs-pastToFemale}: 83 pairs of verbs and its past tense in feminine form. This category is extended from category \emph{past-tense} and is language-specific.
	\item[-] \texttt{verbs-pastToMale}: 83 pairs of verbs and its past tense masculine form. Category is same as past-tense, only its extended variation to be comparable with category \emph{verbs-pastToFemale}.
\end{itemize}

\subsection{Word Similarities Corpora}
For basic comparison with English, we have translated state-of-the-art English word similarity data sets WordSim353 \cite{ws353} and RG65 \cite{RubensteinGoodenough65}. These corpora have 353 (respespective 65) word pairs. Each word pair is manually annotated with similarity. We kept similarities untouched. The words in WordSim353 are assessed on a scale from 0 to 10, in RG65 from 0 to 5.

\section{Distributional Semantic Models}
We experimented with state-of-the-art models used for generating word embeddings. Neural network based models CBOW and Skip-gram from Word2Vec \cite{mikolov2013efficient} tool and tool FastText that promises better score for morphologically rich languages.

\subsection{CBOW} \label{sec:cbow}
CBOW (Continuous Bag-of-Words) \cite{mikolov2013efficient} tries to predict the current word according to the small context window around the word. The architecture is similar to the feed-forward NNLP (Neural Network Language Model) which has been proposed in \cite{bengio2006neural}. The NNLM is computationally expensive between the projection and the hidden layer. Thus, CBOW proposed architecture, where the (non-linear) hidden layer is removed and projection layer is shared between all words. The word order in the context does not influence the projection. This architecture also proved low computational complexity.

\subsection{Skip-gram} \label{sec:skip}
Skip-gram architecture is similar to CBOW. Although instead of predicting the current word based on the context, it tries to predict a words context based on the word itself \cite{mikolov2013distributed}. Thus, intention of the Skip-gram model is to find word patterns that are useful for predicting the surrounding words within a certain range in a sentence. Skip-gram model estimates the syntactic properties of words slightly worse than the CBOW model, but it is much better for modeling the word semantics on English test set \cite{mikolov2013efficient} \cite{mikolov2013distributed}. Training of the Skip-gram model does not involve dense matrix multiplications and that makes training also extremely efficient \cite{mikolov2013distributed}.


\subsection{Fast-Text}
FastText\cite{bojanowski2016enriching} combines concepts of CBOW (resp. Skip-Gram) architectures introduced earlier in Section \ref{sec:cbow} and \ref{sec:skip}. These include representing sentences with bag of words and bag of n-grams, as well as using subword information, and sharing information across classes through a hidden representation.

\section{Experimental Results}

\subsection{Training data}
We trained our models on two datasets in the Croatian language. We made the entire dump of Croatian Wikipedia - dated 08-2017 with approximately 275,000 articles. We have tokenized the text, removed nonalphanumeric tokens and extracted only sentences with at least 5 tokens. Resulting corpus has 92,446,973 tokens. We merged data from Wikipedia with Croatian corpus presented in \cite{vsnajder2013building} that has over 1.2B tokens. Resulting corpus has 1.37B tokens and 56,623,398 sentences. Such corpus has vocabulary of 955,905 words with at least 10 occurences.

For English version of data, we used Wikipedia dump from June 2016. This dump was made of 5,164,793 articles, has 2.2B tokens and vocabulary of xy words. 

We tested analogies and similarity corpora for both languages with most frequent 300,000 words.

\begin{table}[!h]
	\begin{center}
		\begin{tabularx}{\columnwidth}{|l|c|c|}
			
			\hline
			&Vocabulary $tf>10$&Tokens\\
			\hline
			EN corpus& 3,234,907 & 2,201,735,114\\
			HR corpus& 955,905 & 1,370,836,176\\
			\hline
			
		\end{tabularx}
		\caption{Properties of Croatian training data corpus.}
	\end{center}
\end{table}

\begin{table}[!h]
	\begin{center}
		\begin{adjustbox}{max width=0.48\textwidth}
			\begin{tabular}{|l|c|c|c|c|}
				
				\hline
				\bf{Model}&\bf{CBOW}&\bf{Skip-gram}&\bf{fastText-Skip}&\bf{fastText-CBOW}\\
				\hline
				Capital &44.17 & 62.5 & 59.58& 21.25 \\
				Chemical-elements &1.02 & 2.25 & 0.74 & 0.41\\
				City-state & 22.11 & 37.89 & 47.63 & 46.32\\
				City-state-USA & 5.78& 8.23 & 4.30 & 0.37\\
				Country-world & 23.93& 44.49& 40.15 & 7.31\\
				Currency & 4.68& 8.19&6.43 & 0.58\\
				Currency-shortcut & 2.08 & 8.19 & 2.50 & 0.42\\
				EU-cities-states & 21.59 & 41.95 & 42.33  & 6.16\\
				Family & 34.83 & 41.82 & 42.72 & 34.76\\
				\hline
				Jobs & 68.94 & 64.06 & 88.54 & 95.45\\
				Adj-to-adverb & 18.36 & 21.36 & 35.33 &62.01\\
				Opposite & 17.34 & 18.05 & 59.03 & 86.10\\
				Comparative & 34.90 & 33.57 & 43.22 & 41.46\\
				Superlative & 33.22& 27.70 & 40.50 & 51.77\\
				Nationality-man & 17.01& 23.87 & 60.05 & 62.13\\
				Nationality-female & 14.38 & 55.66 & 57.77 & 53.98\\
				Past-tense & 67.31 & 61.03 & 66.67 & 78.21\\
				Plural & 37.12 & 44.65 & 44.24 & 35.10\\
				Nouns-ant. & 12.70 & 10.96 & 10.80 & 21.24\\
				Adjectives-ant. & 13.39 & 13.11 & 18.59 & 12.59\\
				Verbs-antonyms & 9.18 & 6.18 & 7.25& 9.71\\
				Verbs-pastFemale & 60.92 & 19.47 & 71.04 & 80.50\\
				Verbs-pastMale & 66.68 & 62.89 & 76.04 & 85.04\\
				\hline
				\hline
				SEMANTICS\_EN &  73.63 & 83.64 & 68.77 &68.27\\
				SYNTACTIC\_EN & 67.55 & 66.8& 67.94 & 76.58\\
				\hline
				SEMANTICS\_HR & 16.60 & 28.54 & 25.94 & 7.76\\
				SYNTACTIC\_HR & 37.06 & 35.63& 49.60 & 54.56\\
				\hline
				\hline
				\bf ALL\_HR & 32.03 & 33.89 & 43.83 & 43.13\\
				\hline
				
			\end{tabular}
		\end{adjustbox}
	\end{center}
	\caption{\label{tab:resultsHR} Detailed results of Croatian word analogy corpus.}
\end{table}

\begin{table}[!h]
	\begin{center}
		\begin{adjustbox}{max width=0.45\textwidth}
			\begin{tabular}{|l|ccc|}
				\hline
				& \multicolumn{3}{c|}{\bf English}\\
				\bf Models &WordSim353& RG65 & EN-analogies\\
				\hline
				CBOW& 57.94 & 68.69 & 69.98 (44.02)\\
				Skip-gram& 64.73 & 78.27 & 73.57 (46.28)\\
				fastText-Skip& 46.13& 76.31 & 68.27 (42.94)\\
				fastText-CBOW& 44.64& 73.64 & 76.58 (48.17)\\
				\hline
				& \multicolumn{3}{c|}{\bf Croatian}\\
				CBOW& 37.61& 52.01 & 32.03 (19.19)\\
				Skip-gram& 52.16 & 58.47 & 33.89 (20.31)\\
				fastText-Skip& 52.98& 64.31 & 43.83 (25.79)\\
				fastText-CBOW& 30.41 & 51.06 & 43.14 (25.79)\\
				\hline
				
			\end{tabular}
		\end{adjustbox}
		\caption{Comparison with English models. Measurement in brackets gives the results including OOV questions.}
	\end{center}
\end{table}

In total we tested on 68,986 out of 115,085 questions, it means that almost 40\% question was unknown by model. All question conatined OOV words were discarded from testing process. We tested Semantic group on 16,968 known questions and part of corpus testing syntactic properties was measured on 52,018 questions. 

Only 10 out of 353 question was unknown for \emph{WordSim353} corpus and all 65 questions of \emph{RG65} were in vocabulary. Unknown words in \emph{WordSim353} were represented as word vector averaged from 10 least common words in vocabulary.\\

Semantic tests gives overall poor performance on all tested models, as we can see in Table \ref{tab:resultsHR}, the opposite is true for English, where semantic tests gives usually similar score as syntactic tests. This behavior we already saw on Czech corpus presented in \cite{DBLPSvobodaB16}. It seems that free word order and other properties of highly inflected languages from Slavic family have a big impact on the performance of current state-of-the-art word embeddings methods.\\
From results of \emph{City-state} and \emph{City-state-USA} category it can be seen that knowledge of the topic in training data has significant impact on performance of a model. We wanted to show differences between two similar categories in case we have an insufficient amount of training data covering a particular topic. Category \emph{City-state} is showing that model is able to carry such knowledge - if the topic is sufficiently represented in a training data, the model is able to carry this type of information. This behavior is seen in regions from Croatia mentioned in many articles on Croatian Wikipedia, but this was not a case with states from USA. All questions of \emph{City-state} were covered, but only around 50\% of questions in category \emph{City-state-USA} were in vocabulary. On categories \emph{Country-world} and \emph{EU-cities-states} it can be seen that there is no difference between knowledge about states and main cities from EU again state-city pairs from all over the world. Another very poor performance gives group \emph{Currency}, but this group is usually weak across all languages and shows the weaknesses of the model.\\

Syntactic tests gives better performance than tests oriented to semantic, but they still have significantly worse performance rather than on English. This part of corpus includes language-specific group of tests - such as \emph{Verbs-pastMale/Female}, \emph{Nationality-man/female}. Simple \emph{Past-tense} tests gives surprisingly high score - similarly it was also with Czech language in \cite{DBLPSvobodaB16}. We could say, that languages from Slavic family tends to have easier patterns for past tense. From language-specific groups we see that slightly better score is given in categories with word pairs in the masculine form, these results also corresponds with the fact that there are more articles written in masculine form in the training data.

\section{Conclusion}
In this paper evaluation of Croatian word embeddings are performed. New corpus from the original Word2Vec is derived. Additionally, some of the specific linguistic aspects of the Croatian language was added.  Two popular word representation models were compared, Word2Vec and fastText. Models have been trained on a new robust Croatian analogy corpus. WordSim353 and RG65 corpuses were translated from English to Croatian, in order to perform basic semantic measurements. Results show that models are able to create meaningful word representation. 

However, it is important to note that this paper presents the first comparative study of word embeddings for Croatian and English, and therefore, new insights for NLP community according to the behavior of the Croatian word embeddings. Croatian belongs to the group of Slavic languages and has only preliminary and basic knowledge insights from word embeddings. In addition, another contribution of this work is certainly new data sets for the Croatian language, which are publicly available from: \url{https://github.com/Svobikl/cr-analogy} These are also the first parallel English-Croatian word embeddings datasets.

Finally, we can figure out from experiments that models for Croatian does not achieve such good results as for English. According to \cite{DBLPSvobodaB16}, this is also true for the Czech language, another one from Slavic language family. Following this, we would like to point out that future research should be focused on model improvements for Slavic languages. The difference in English and Slavic language morphology is huge. Compared to the Croatian language, English language morphology is considerably poor. Croatian is a highly inflected language with mostly free word ordering in sentence structure, unlike English, which is inflectional language and has a strict word ordering in a sentence. These differences are reflected in the results of embeddings modeling. Models give good approximations to English, they are better tailored to the English language morphology and better match the structure of such a language. In future research, it would be worth to explore which Slavic languages specificities would be advisable to incorporate into models, in order to achieve better modeling of complex morphological structures. On the other hand, corpora preprocessing which simplifies morphological variations, such as stemming or lemmatization procedures, could also have an effect on word embeddings and should be one of the future research directions.

For a future work we would like to further investigate properties of other models for word embeddings and try to use external sources of information (such as part-of-speech tags, referenced information on Wikipedia, etc.) and experimenting with tree structure of sentence during training process.

\section{Acknowledgements}

This work was supported by the project LO1506 of the Czech Ministry of Education, Youth and Sports and by Grant No. SGS-2016-018 Data and Software Engineering for Advanced Applications.

\section{Bibliographical References}
\label{main:ref}

\bibliographystyle{lrec}
\bibliography{cranalogy}

\end{document}